\documentclass[letterpaper, 10 pt, journal, twoside]{IEEEtran}
\usepackage{url}

\usepackage{math}
\usepackage[all]{nowidow}

\newcommand{\rev}[1]{\textcolor{black}{#1}}
\long\def\invis#1{}

\ifCLASSINFOpdf
\else
\fi
\begin{document}
%
\title{\LARGE \bf AdaptiveON: Adaptive Outdoor Local Navigation Method for Stable and Reliable Actions}
%
%
%

\author{Jing Liang, Kasun Weerakoon, Tianrui Guan, Nare Karapetyan and Dinesh Manocha
\thanks{The authors are with the University of Maryland, College Park, MD 20742, USA. {\tt\small [jingl, kasunw, rayguan, knare, dmanocha]@umd.edu}}
}
%
%

\markboth{IEEE Robotics and Automation Letters. Preprint Version. Accepted December, 2022}
{Liang \MakeLowercase{\textit{et al.}}: AdaptiveON: Adaptive Outdoor Local Navigation Method for Stable and Reliable Actions} 

%



\maketitle

\begin{abstract}

We present a novel outdoor navigation algorithm to generate stable and efficient actions to navigate a robot to reach a goal. We use a multi-stage training pipeline and show that our approach produces policies that result in stable and reliable robot navigation on complex terrains. Based on the Proximal Policy Optimization (PPO) algorithm, we developed a novel method to achieve multiple capabilities for outdoor local navigation tasks, namely alleviating the robot’s drifting, keeping the robot stable on bumpy terrains, avoiding climbing on hills with steep elevation changes, and avoiding collisions. Our training process mitigates the reality (sim-to-real) gap by introducing generalized environmental and robotic parameters and training with rich features captured from light detection and ranging (Lidar) sensor in a high-fidelity Unity simulator. We evaluate our method in both simulation and real-world environments using Clearpath Husky and Jackal robots. Further, we compare our method against the state-of-the-art approaches and observe that, in the real world, our method improves stability by at least $30.7\%$ on uneven terrains, reduces drifting by $8.08\%$, and decreases the elevation changes by $14.75\%$. 
\end{abstract}

\begin{IEEEkeywords}
Autonomous Vehicle Navigation, Motion Control, Reinforcement Learning.
\end{IEEEkeywords}

Code: \url{https://github.com/jingGM/adaptiveON.git}

Project Page: \url{https://gamma.umd.edu/adaptiveon}

\section{INTRODUCTION}
\label{section:introduction}
The mobile robot navigation problem in indoor and outdoor scenes has been studied extensively.\invis{It is an important field of research, especially with the rapid growth of autonomous vehicles. } Nevertheless, outdoor navigation remains an unresolved challenge when dealing with unstructured terrains and environmental uncertainties~\cite{kahn2021badgr,nagariya2020iterative}. 



Robot navigation in complex terrains is severely affected by different characteristics of the environment~\cite{terp,guan2021ganav}. The texture of the ground surface may affect the friction between the robot and the terrain. For example, mud, leaves, or water may cause robots to drift~\cite{wellhausen2019should,wapnick2021trajectory}.  

For stable navigation, along with collision avoidance, reducing vibration and preventing high elevation changes in motion are important objectives that will help prevent motor damage and ensure accurate motions~\cite{terp,yihuan2011motion}. Several works use segmentation methods to classify the uneven terrains based on traversability~\cite{guan2021ganav,hirose2018gonet,viswanath2021offseg,eder2021autonomous}. 
However, segmentation suffers from the noise and complexity of real-world features; for different robots, the traversability of the same area could be different based on their dynamics.\invis{For high hills, \cite{terp} tries to generate way points to produce a trajectory with lower elevation changes, but on the action level it is hard to maintain low elevation change at each step while the robot moves between two way points.}


\begin{figure}
\centering
\minipage{0.24\textwidth}
  \includegraphics[width=\linewidth,height=0.7\linewidth]{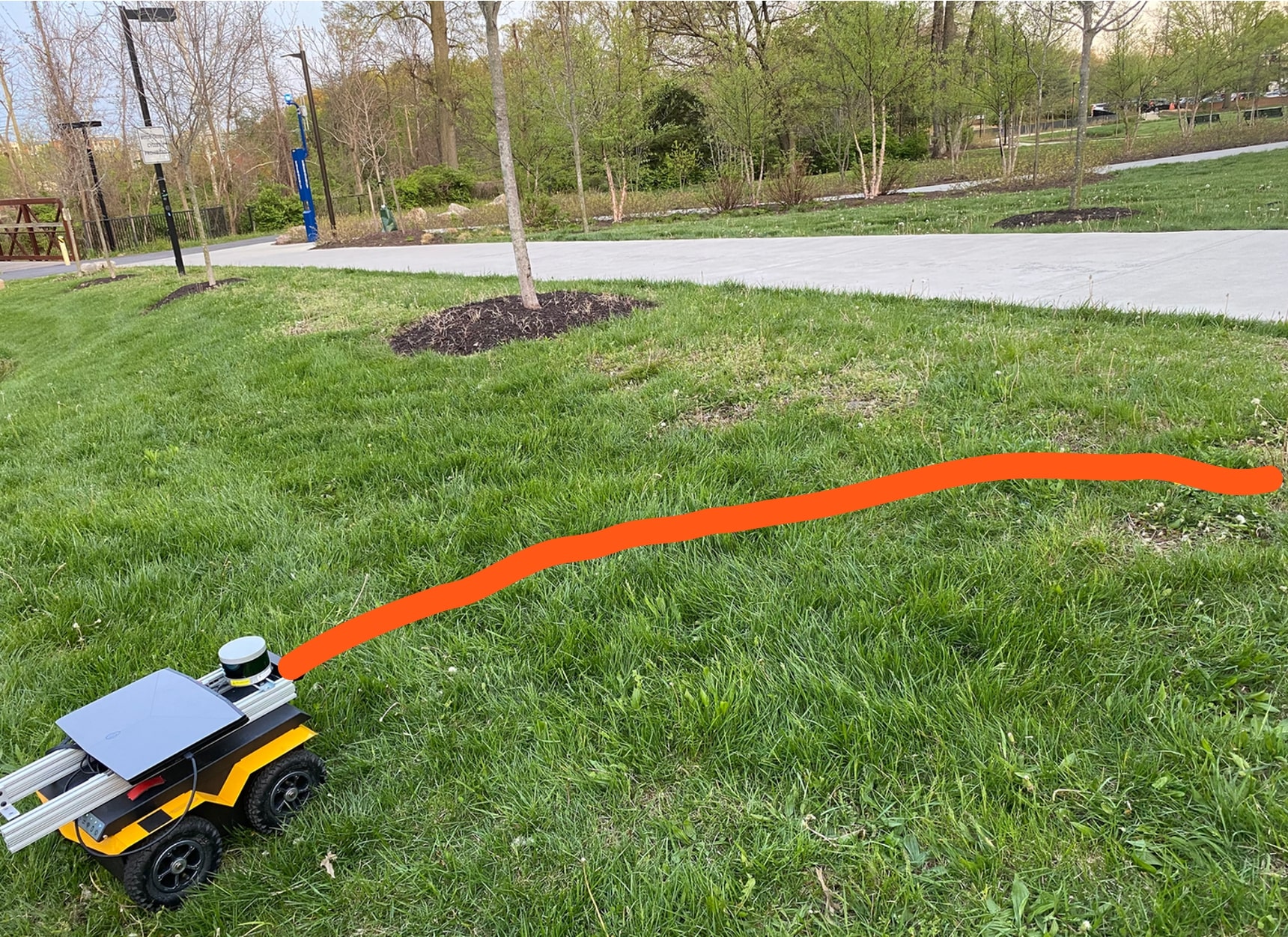} 
\endminipage\hfill%
\minipage{0.24\textwidth}%
  \includegraphics[width=\linewidth,height=0.7\linewidth]{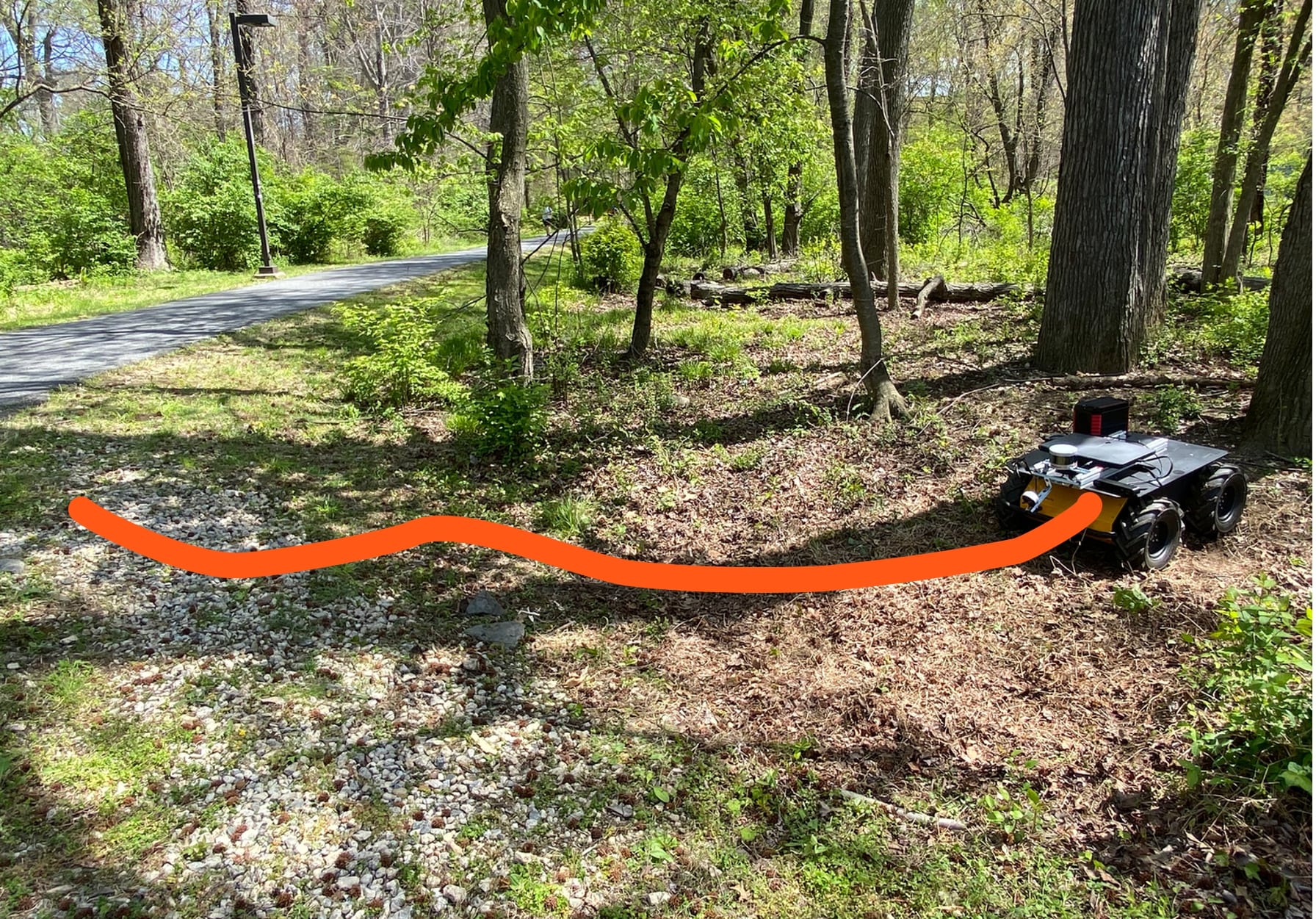}
\endminipage
    \caption{
    Trajectories generated by AdaptiveON (ours) when navigating Clearpath Jackal and Husky robots on complex outdoor terrains: Our method ensures stable navigation by reducing vibrations of the Jackal robot on a slope with uneven texture (left); AdaptiveON provides reliable motion control along the trajectories to prevent the Husky from drifting on slippery terrains, including weathered leaves, small rocks, and grass (right).
    }
    \label{fig:frontpage}
\end{figure}


Simulations provide the means to perform extensive and safe training for navigation tasks. Nevertheless, the policies learned in these environments are always bound by the characteristics of the simulations and hence suffer when generalized to the real world. The sim-to-real problem or so-called ``reality gap" arises when many uncertainties are included: the physical properties of the robot, the world model differences, and the dynamics of the simulated world. This means that better navigation requires real-world training. However, training in the real world is not always feasible, especially when dealing with expensive robots with stability constraints~\cite{hwangbo2019learning,kumar2021rma}. Some of the approaches for closing the reality gap train the RL system in simulation and then use sim-to-real methods~\cite{tobin2017domain}. Alternative strategies aim to develop adaptive systems that do not require real-world training~\cite{peng2020learning,lee2020learning,hu2021sim}.

\textbf{Main Contributions: } We present AdaptiveON, a novel approach with a multi-stage pipeline based on the PPO algorithm for training robots that avoid obstacles and perform stable and reliable navigation on hills, uneven and slippery areas in complex outdoor scenarios. Detailed contributions are as follows: 
\begin{enumerate}
    \item \rev{We introduce the "Adaptive Model" to embed the recent history of the robot's status, which helps navigation on terrains with different properties.} Incorporating it with perception models, we adapt the robot to different terrains and train the robot with reliable navigation. For instance, moving on slippery terrains with less drift, moving on hills with low elevation changes on each action, and avoiding collisions with static obstacles. We show that our method has less drifting than other state-of-the-art approaches.
    
    \item Along with the encoded information of the history of the robot's states and actions, we also introduce a module for stability, “Stable Model,” to adaptively control the thresholds of the robot's velocity on different types of terrains. The model can keep the robot at a relatively low speed on bumpy surfaces and make the robot stable.
    
    \item We alleviate the reality gap by designing training scenarios in the RaiSim simulator~\cite{raisim} with different environmental, mechanical, and control parameters that could affect the robots' motions, e.g., frictions, unevenness, PID parameters, and weights. Additionally, we fine-tune the robot's perception in the Unity simulator with rich features.
\end{enumerate}

The rest of this paper is organized as follows. First, an overview of related work is presented in Section~\ref{section:related_work}. In Section~\ref{section:approach}, we give a formal definition of the problem and present details of the method. We report the experimental results of our method compared to state-of-the-art approaches in Section~\ref{section:experiments}. Finally, conclusions are drawn in Section~\ref{section:conclusions}. 


\section{Related Work}
\label{section:related_work}

Approaches for addressing the robot navigation problem in outdoor environments vary from those specific to indoor environments primarily due to the complexity of the outdoor terrains. When the state of the environment and the kinematics of the system are known, some classical methods~\cite{van2011reciprocal,fox1997dynamic,egograph} have proven to be effective in tackling navigation; nevertheless, when the environment is challenging and changes are stochastic, learning-based approaches are more practical~\cite{kahn2017uncertainty,terp,ibarz2021train,weerakoon2022graspe}.  


\subsection{Learning-based Navigation}
\label{subsection:Learning_Based_Navigation}

Deep Reinforcement Learning (DRL) has been widely used in complex control problems under uncertainties. 
Brunner et al. \cite{brunner2018teaching} used DRL for indoor navigation to drive out of a random maze by learning to read a global map and finding the shortest path. However, retraining is required after changing targets when the navigation has been designed for specific tasks. Staroverov et al. \cite{staroverov2020real} proposed hierarchical DRL with classical methods integrated into the model. The navigation model presented by \cite{pfeiffer2018reinforced} is an end-to-end neural network trained with a combination of expert demonstrations, imitation learning, and reinforcement learning. 

DRL-based navigation methods show promise, but mostly in simulated environments. These methods are often difficult to execute in the real world due to underlying challenges such as sample efficiency or data synchronization \cite{ibarz2021train, weerakoon2022htron}. On the other hand, some of the approaches require real-world training or data collection, which might be non-trivial. BADGR~\cite{kahn2021badgr} uses a self-supervised learning approach, which needs to experience real-world events such as colliding with obstacles to learn its policy.
 
In our prior work, we used learning-based approaches to navigate in dense crowds~\cite{densecavoid,roth2021xai,crowdsteer}, in scenarios with fast moving obstacles~\cite{ofvo}, or in outdoor terrains by selecting locations with low elevation changes to travel through~\cite{terp}. We also used segmentation methods for identifying navigable regions \cite{guan2021ganav}. These works are limited by the predefined characteristics of the traversability of a region associated with the specific type of robot. In this work, we introduce an adaptive model that will allow us to alleviate those dependencies.

\subsection{Sim-to-Real}
\label{subsection:Sim_to_Real}
The problem of transferring from simulation to the real world is called the reality gap. Learning in the real world is challenging because robots can be damaged when interacting with the world. Hwangbo et al.~\cite{hwangbo2019learning} proposed a method based on the Trust Region Optimization Policy (TRPO) that learns gaits of the robot. Their model produced the fastest moving and most agile actions at the time for legged robots. The model has been deployed on the ANYmal robot, which successfully completed a 2.2km length trail up a mountain and back. CrowdSteer~\cite{crowdsteer} used PPO to train the policy in different scenarios for collision avoidance and achieved relatively good performance.

Outdoor navigation for legged robots based on adaptation from simulation has been proposed by~\cite{kumar2021rma}. Their approach, Rapid Motor Adaptation (RMA), consists of two components: a base policy and an adaptation module. It is trained completely in simulation without using any domain knowledge and deployed on the robot without any fine tuning. This ensures that, for complex systems such as legged robots, no additional real-world training is required. However, the objective of this work is to generate only stable motions without considering target-approaching objectives.


End-to-end DRL strategy \cite{zhang2018robot} incorporates an elevation map, depth images, and orientation details to train a DRL policy to navigate in uneven terrains. In \cite{josef2020deep}, a DRL method for local planning in unknown rough terrain with zero-range to local-range sensing is proposed.  The method utilizes ground elevation data, self-attention modules, the geometric transformation between two successive timesteps, and the corresponding action to navigate on surfaces with different levels of friction and elevation. However, these methods are trained and tested only in simulated environments.

\section{APPROACH}
\label{section:approach}
 
In this section, we formally define the problem,
introduce the architecture of the neural networks, and discuss the training details.
\subsection{Problem Formulation}

In this project we use Lidar, odometer and IMU as primary sensors to perceive the environment and robot's status. We assume that the system states can be estimated from the observations. 
Thus, we formulate the outdoor navigation problem as Markov Decision Process (MDP) problem:
    \[\cp = (\cs, \ca, \ct, \crr, \gamma ), \] 
    where $\cs$ is the state space, and we have each state $\s=\left(\o_l, \o_s \right) \in \cs$, which contains the lidar observation, $\o_l$, and the current status of the robot, $\o_s$, perceived by IMU and odometer. $\ca\in \RR^2$ is the action space, and each action $\a=\left(v,\omega\right) \in \ca$ has linear and angular velocities, $v\in [0,1](m/s)$ and $\omega\in [-1,1](r/s)$, respectively. Each action is chosen by the end-to-end network $\a=\pi_s(\s)$.
    $\ct$ is the transitional probability from the current state and action to the next state, and $P(\s_{t+1}|\s_t,\a_t)$ denotes its probability.
    $\crr: \cs \times \ca \rightarrow \RR$ represents the mapping from states and actions to rewards. In each time step, we have the reward $r(\s_t, \a_t)\in\crr$, where $\s_t$ and $\a_t$ are the state and action at the time step t, and the reward is observed from the environment/simulator.
   The discount factor $\gamma$ is set to $0.97$. Denote $\pi_\theta(\s_t)$ as the policy, which represents the entire network. $\theta$ represents the parameters of the networks.
    
    We use the PPO reinforcement learning algorithm to solve this MDP problem, and the objective of the algorithm is to find the $\pi_{\theta^*}(\s_t)$ that achieves the highest rewards.
    We use different reward component functions for different navigation functionalities, including collision avoidance, stability maintenance, drift prevention, and low elevation changes.

The \textit{\textbf{stability}} objective in this work is to minimize the vibrations while moving. The vibration $vb$ is defined as the sum of angular velocities of roll and pitch angles $\omega_r$ and $\omega_p$, respectively. 
\begin{align}
    vb = \abs{\omega_r}+\abs{ \omega_p}
    \label{eqn:single_stability}
\end{align}

We call the navigation \textit{\textbf{reliable}} if the robot can successfully reach the goal while avoiding obstacles, achieves lower elevation changes, and has less drifting. To be more specific, the drifting issue is always caused by low friction, i.e., given a certain velocity the robot may not move as predicted.

\subsection{Architecture}
\label{sec:architecture}

The architecture of the proposed approach consists of several components, each designed for one capability of the robot.
In addition to collision avoidance, our objectives are to prevent drifting, maintain stable navigation on uneven terrain, and keep changes in the elevation of the robot small in each time-step. Accordingly, there are 4 models embedding the information required by the four functional targets and a single Policy Model, which takes this information and generates actions for the robot. The complete architecture of our proposed method consists of perception models (Lidar Model and Elevation-map Model), Adaptive Model, Stable Model and Policy Model (see Figure~\ref{fig:archit} for the architecture and details in Supplement of \cite{adaptiveon}).

\begin{figure}[t]
\centering
\includegraphics[width=\linewidth]{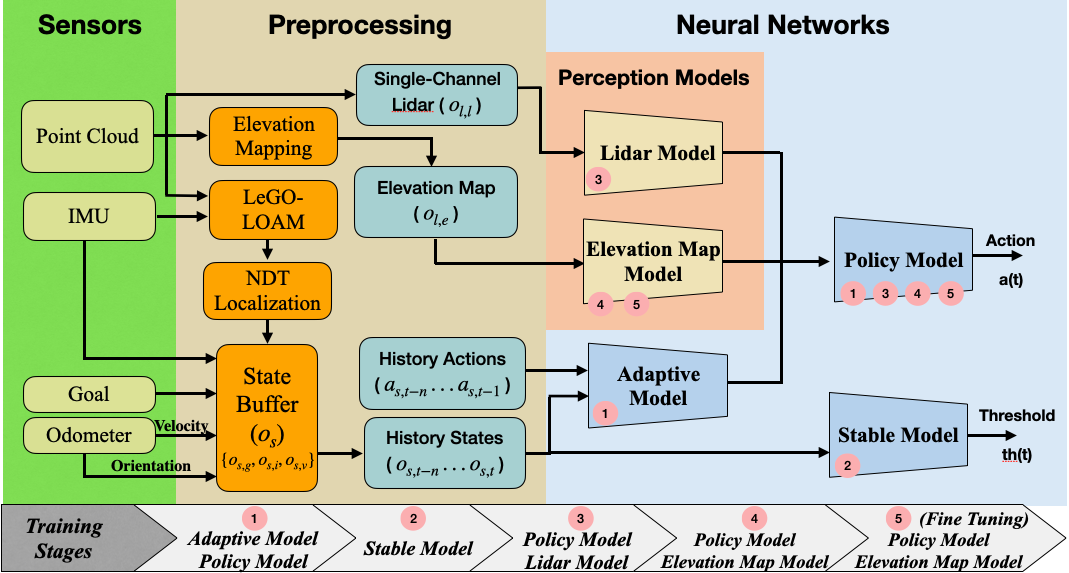}
\caption{System Architecture consisting of 5 models: Perception Models (Lidar Model and Elevation Map Model); Adaptive Model; Policy Model that takes the output from all the other models and generates actions resulting in reliable navigation; and Stable Model, which generates an action threshold to ensure stability. The training pipeline, on the bottom of the figure, is explained in Section \ref{sec:pipeline}.}
\label{fig:archit}
\end{figure}

Lidar is widely used in navigation tasks~\cite{crowdsteer,densecavoid,ofvo} due to its accurate distance estimation capabilities. In this work, we use a 16-Channel Velodyne Lidar as the primary perception sensor. Our perception models include the Lidar and the Elevation-map Models. These perception models are used to extract the features of the nearby environment of the robot. We use a pointcloud-based elevation map~\cite{elemap} to detect the terrains in the robot's vicinity. \invis{One of the shortcomings of the elevation map is that it cannot detect obstacles accurately} However, the elevation map cannot identify the obstacles accurately. Therefore, we use the middle channel of the raw Lidar data to extract the nearby obstacles, as shown in Figure \ref{fig:archit}. 
The Lidar Model is
composed of a sequence of convolutional layers and fully connected layers, and the Elevation-Map Model is composed of 2-D convolutional layers and fully connected layers. The output of the Lidar Model is $\o_{l,l}$, and the output of the Elevation-Map Model is $\o_{l,e}$; hence, the Lidar observation is $\o_l=(\o_{l,l}, \o_{l,e})$. The input and network details are shown in the supplement \cite{adaptiveon}. 

Inspired by RMA \cite{kumar2021rma}, we also use the previous consecutive frames of the robot's observed states and actions to \rev{estimate the robot's current dynamic status in motion.
Those states and actions} are states $\o_{s,t-n}...\o_{s,t}$ and actions $\a_{t-n}...\a_{t-1}$. Each state $\o_{s}=\{\o_{s,g}, \o_{s,i}, \o_{s,v}\}$ contains the information of the relative goal position $\o_{s,g}$ (detected from the positioning model in Section \ref{section:experiments}), filtered IMU data $\o_{s,i}\in \RR^6$ from the IMU sensor, and the robot's current velocity $\o_{s,v}$ from the odometry sensor. The goal observation $\o_{s,g} = (d_g, a_g, h_g)$ comprises the distance, angle, and height from the robot's current position to the goal position. $\o_{s,i}$ contains the linear accelerations $(a_x, a_y, a_z)$ \invis{w.r.t. the robot's coordinate frame}, angular velocities $(\omega_r, \omega_p, \omega_y)$ of roll, pitch, and yaw rotations, and the roll, pitch, and yaw angles $(r, p, y)$. Further, $\o_{s,v}$ includes the linear and angular velocities ($v,\omega$) of the robot obtained from the odometer (see Figure \ref{fig:archit}). Finally, we concatenate relative goal position, IMU, and Odometer data together as a state. The details of the Adaptive Model are in the Supplement section of \cite{adaptiveon}.


As shown in Figure \ref{fig:archit}, there are two output models: the Policy Model and the Stable Model. The Policy Model takes the embedded values from perception models and the Adaptive Model to generate actions, $\a_p$, to move the robot, and the Stable Model takes the historic state information, $\o_{s,t-n}...\o_{s,t}$, to generate the threshold, $\t\h$, for the actions of the robots. These two models are all composed of fully connected layers. We use the final clipped actions, $\a=\min(\t\h, \a_p)$, to drive the robot. The details of the Output Models are in the Supplement section in \cite{adaptiveon}.




 
\subsection{Training}






\subsubsection{Training Objectives}

\begin{figure*}[t]
\centering
\includegraphics[width=0.9\linewidth]{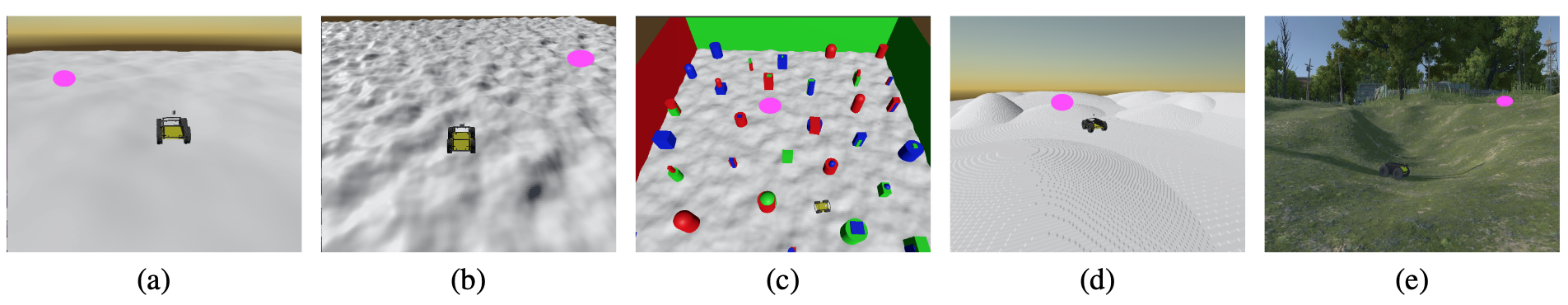}
 
\caption{Different Scenarios in simulation used for training different functionalities of robots. The purple circles denote the given goals. (a) has different environmental, mechanical, and controlling parameters for the robot. This type of scenario trains the policy to adapt to different outdoor conditions and prevent drift. (b) has high unevenness and is used to train the Stable Model to reduce vibration. (c) is used to train the robot to avoid collisions with obstacles. (d) contains hills with different heights and gradients, and it is used to train the robot to move in lower elevation changes in each step. (e) includes most of the properties of the previous scenarios and has rich features to fine-tune the perception networks.}
\label{fig:scenarios}
\end{figure*}

By training different models in different scenarios, the policy can obtain different functionalities: preventing drifts, running slower on uneven terrain, avoiding collisions, and having small elevation changes when it runs on hills. Since the Raisim simulator \cite{raisim} provides better tools to customize terrain features, we train all the basic functionalities in Raisim using the scenarios shown in Figure  \ref{fig:scenarios}.

The other objective of the training is to alleviate the differences between the simulation and the real world. To this end, we focus on two aspects: First, in real-world outdoor scenarios, terrains can have different levels of frictions, unevenness and restitution, and the robots can have different weights, PID parameters, and joint numbers. We designed Raisim simulation environments with different values of these parameters (discussed in Section \ref{sec:pipeline}) and trained the Adaptive Model to take the robot's observed states in the scenarios with these different properties. Second, the real world has richer features than the Raisim simulator w.r.t. Lidar data. Hence, we use a high-fidelity Unity simulator, which contains significantly rich features, to fine-tune the parameters of the perception models.
    
    
    
    

\subsubsection{Reward functions}
\invis{In this section, we introduce the reward function of the whole training process.} The general reward function is:
\begin{equation}
    r = \left\{\begin{matrix}
                 r_t  &\text{if terminated,}\\ 
                \beta_g r_g + \beta_d r_d + \beta_s r_s + \beta_c r_c + \beta_e r_e &\text{otherwise}, \end{matrix}\right.
\label{eq:all_rewards}
\end{equation}
where the values $\beta_g, \beta_d, \beta_s, \beta_c, \beta_e$ are coefficients of the rewards. During different training stages, we use higher weights to the corresponding rewards and lower otherwise (even 0). In this work, we only consider (0,1) for the weights, which are discussed in Training Pipeline Section~\ref{sec:pipeline}. To efficiently train the different models with different functionalities and training stages, the rewards are weighted according to those functionalities.

\begin{enumerate}
 \itemsep-1mm 
    \item Distance to Goal $r_g$: This reward encourages the robot to move towards the target and penalizes it for deviating. $d_{g, \text{last}}$ and $d_{g, \text{now}}$ are the distances between the robot and the goal in the last and the current time step, respectively:
    \begin{equation}
        r_g = d_{g, \text{last}} - d_{g, \text{now}}.
    \end{equation}

    \item Drift $r_d$: This novel reward function penalizes drifting on slippery terrains. $\p_a$ is the position of the robot in the next time step calculated by the given action, and $\p_r$ is the real position in the next time step:
    \begin{equation}
        r_d = - \norm{\p_a - \p_r}.
    \end{equation}

    \item Stability $r_s$ : This reward encourages the robot to run with more stability, which is $vb$ as defined in Equation \ref{eqn:single_stability}. The $vb_{th}$ is the stability threshold, which we consider as 0.5 during the experiments:
    \begin{align}
        r_s &= \left\{\begin{matrix}
                0  , & vb \leq vb_{th}\\ 
                vb_{th}-vb , & vb>vb_{th},
                \end{matrix}\right.
    \end{align}
 
    \item Obstacles $r_c$: This reward penalizes the robot when it runs close to obstacles. $c_{th}=1.5 \times r_{\text{rob}}$ is the distance threshold to obstacles, where $r_{\text{rob}}$ is the radius of the robot. $d_c = \min{(\d_o)}$ is the nearest distance to all the nearby obstacles. Here, $\d_o$ is the single-channel Lidar data, which is a vector of sampled distances:
    \begin{align}
        r_c &= \left\{\begin{matrix}
                0  , & d_c > c_{th}\\ 
                \exp(d_c-c_{th}) , & d_c\leq c_{th},
                \end{matrix}\right. 
    \end{align}
 
    \item Elevation changes $r_e$: This reward encourages the robot to move to areas with lower elevation changes. The elevation change is represented by the gradient of the elevation map.  $\g_e$ represents the gradient vector along the heading direction of the robot:
    \begin{align}
        r_e = \norm{\g_e},
    \end{align}
 
    \item Termination rewards $r_t$: This reward is triggered during specific events such as: robots reaching the goals $r_\text{success}$, collisions $r_\text{collision}$ , or robot flip-overs $r_\text{flips}$:
    \begin{align}
        r_t = \left\{\begin{matrix}
                 r_\text{success}=100  &\text{If reaches the goal,}\\ 
                r_\text{collision}=-100 &\text{If in collision,} \\
                r_\text{flips}=-100 &\text{If flip-over},
                \end{matrix}\right.
    \end{align}
\end{enumerate}

\subsubsection{Training Pipeline}
\label{sec:pipeline}
\invis{The following is the sequence of each training stage.} We train the functionalities consecutively in the scenarios listed below.
While training a specific functionality, we only make the parameters of relative models derivable. Other models are fixed, and the untrained models are initialized with 0 to prevent disturbance to the current training stage. 
 Before the Uneven Terrain, we use the maximum threshold $th=[1m/s, 1r/s]$ to substitute the output of the Stable Model.

\begin{enumerate}
    \item Adaptive Scenarios (Figure \ref{fig:scenarios} (a)): 
    This scenario has different environment parameters, including different levels of friction (coefficient in [0.3-5.0]) and unevenness (different densities of bumps and height changes in the range [-1m, 1m] of the terrain). We also employ robots with different joints (2 and 4 joints) and control (PD) parameters to joints, where the PD values are the same for all joints in one robot. This stage trains the policy with two functionalities: 1. To make the policy adaptive to different environmental, robotic, and controlling properties; 2. To make the robot able to reach the target. In this stage we only train the Policy and the Adaptive Models and keep other models fixed. For the reward function, we only use $r_g$ and $r_d$ in this stage, and others are weighted by 0.
    
    \item Uneven Terrain (Figure \ref{fig:scenarios} (b)): The scenarios have different levels of unevenness generated by Perlin noise, and the heights of the peaks vary from 0 meters to 1 meter. The unevenness causes vibrations when robots navigate on the terrain. The training in this type of scenario makes the Stable Model generate appropriate velocity thresholds to keep the robot stable while it is running on the uneven terrain. During this training stage, we fix the perception models, the Adaptive Model, and the Policy Model. Hence, we use $r_s$ and $r_g$ in the reward function only to train the threshold of the actions; other rewards are weighted by 0.
    
    \item Scenarios with Obstacles (Figure \ref{fig:scenarios} (c)): This scenario has dense static obstacles (two to three meters apart). The obstacles consist of general features (corners, curves, lines) w.r.t. Lidar data. Hence, training in this stage encodes the distance information of the nearby obstacles by the Lidar Model. The input of the Lidar Model is the 1-channel Lidar data. In this stage, we only train the Lidar Model, while the Policy Model and other models are fixed. The reward function depends on $r_g$ and $r_c$; other rewards are weighted by 0.
    
    \item Scenarios with different hills (Figure \ref{fig:scenarios} (d)): The hills have different heights and gradients, which are  generated randomly within the ranges $[0.1m, 2m]$ and $[10^\circ, 50^\circ]$, respectively. In some regions with large gradients, the robot will flip over if the pitch or roll angles are significantly high. Hence, this stage is used to train the Elevation-map Model to encode the elevation information and produce a policy that can run the robot with a smaller change in elevation in each step. To this end, we use $r_g$ and $r_h$ in the reward function, and other reward items are weighted by 0.
    
    \item Unity Scenarios (Figure \ref{fig:scenarios} (e)): The unity simulator consists with a feature-rich outdoor environment. Lidar and elevation data, specifically, are more complex and accurate than the Raisim simulator. We fine-tune the perception models in the Unity simulator to equip them with better encoding of the environment, i.e., we keep non-zero weights only for $r_g$ and $r_h$.
\end{enumerate}

During training, except for the elevation map, all other inputs are directly from sensors. For elevation maps, we use the elevation mapping package~\cite{elemap}. However, the generated elevation map cannot cover all the areas near the robot without performing a complete exploration of the environment. Further, the elevation map is processed at a relatively slower rate. We alleviate the issues with the following two steps:
\begin{enumerate}
    \item We use the linear interpolation method to fill the unknown regions in the elevation map and solve the sparsity issue of the generated elevation map.
    
    \item To mitigate the slower processing rate, we first use the ground truth elevation map during the training in the Raisim simulator to let the robot learn how to achieve the goal by running on the areas with lower elevation changes. Then, we train the model in the Unity simulator with richer environmental features and a real elevation map.
    \invis{\textcolor{red}{I am not sure if 1 and 2 should be included here, this is implementation detail}}
\end{enumerate}

For each stage of training, we use the trained model from its last training stage, and the training curves are shown in the Supplement in \cite{adaptiveon}.

\section{Benchmarks}
\label{section:experiments}
To show the benefits of our approach in terms of stability (less vibration) and reliability (low elevation changes, less drifting, fewer collisions), we perform experiments in simulation and in real-world scenarios to compare against Terp~\cite{terp}, CrowdSteer~\cite{crowdsteer}, and Ego-graph~\cite{egograph} algorithms, which are discussed in Sections \ref{section:introduction} and \ref{section:related_work}, by measuring the following metrics: trajectory length, vibration, success rate, elevation changes, and drifts. Additionally, we compare the versions before/after different training stages in simulation to show the benefits of each model, summarized in Table \ref{tab:simulation_results}.


Ego-graph uses an elevation map as the cost map to choose the waypoints and moves the robot to follow the waypoints. Similar to Ego-Graph, Terp also generates waypoints in the elevation map, but it uses the DWA~\cite{dwa} motion planning algorithm for navigation. CrowdSteer is an end-to-end motion planning method with Lidar and IMU data as its inputs. 

\begin{table}[h!]
\vspace{2mm}
\resizebox{\linewidth}{!}{
\begin{tabular}{|c|c|c|c|c|c|c|c|} 
\hline
\multirow{3}{*}{Scenarios}  & \multirow{3}{*}{Methods} & Trajectory & Time & Success & Vibration & Elevation & Drift\\ [0.5ex]
 &  &  Length & & Rate  & (r/step) &  Changes & (r/step) \\ [0.5ex]
 &  &  (m) & (s)&   &   &  (cm/step) & $\times$ 100 \\ [0.5ex]
\hline
\multirow{3}{*}{Uneven} 
& AdaptiveON              & 7.25 & 35.5 &1.0 & \textbf{0.79} & 0.19 & \textbf{4.90} \\
& AdaptiveON Before Stage 2   & 6.83 & 7.80 &1.0 & 0.85 & 0.47 & 6.94 \\
\multirow{3}{*}{Terrain} 
& Base Policy & 6.30 & 36.7 &1.0 & 0.93 & 0.19 & 5.12 \\
& Ego-graph               & 7.31 & 9.05 &1.0 & 0.84 & 0.53 & 11.09 \\
& CrowdSteer              & 7.11 & 8.00 &1.0 & 0.81 & 0.45 & 5.46 \\
& Terp                    & 7.21 & 14.2 &1.0 & 0.85 & 0.30 & 8.23 \\
\hline

\multirow{2}{*}{Static} 
& AdaptiveON              & 9.5 & 23 & 0.9 & 0.9 & 0.34 &  3.8\\
& AdaptiveON Before Stage 3  & - & - & 0 & - & - & - \\
\multirow{2}{*}{Obstacles}
& Ego-graph               & - & - &- & - & - & - \\
& CrowdSteer              & 7.82& 9.2 & \textbf{1.0} & 0.93 & 0.45 & 8.19 \\
& Terp                    & 8.73 & 14.9 & 0.7 & 0.93 & 0.31 & 3.99 \\
\hline

\multirow{4}{*}{Hills} 
& AdaptiveON              & 11.50 & 40.7 &1.0 & 0.7 & \textbf{0.12} & 1.5\\
& AdaptiveON Before Stage 4  & 8.51 & 43.3 &1.0 & 0.81 & 0.15 & 1.34\\
& Ego-graph              & 8.37 & 13.2 &1.0 & 1.5 & 0.20 & 7.73\\
& CrowdSteer             & 10.9 & 12.3 &0.9 & 1.4 & 0.21 & 3.22 \\
& Terp                   & 7.68 & 18.1 &1.0 & 0.7 & 0.18 & 9.75\\
\hline
\end{tabular}
}
\caption{\label{tab:simulation_results} Simulation results: The lower the Vibration, Elevation Changes, and Drift values, the better (more reliable) the performance. AdaptiveON shows better stability on uneven terrain, ensures less drifting and smaller elevation changes, and is more reliable than other methods. (Note Ego-graph and the policy without training in Scenario c (Stage 3), in Figure \ref{fig:scenarios}, do not avoid obstacles, and thus they are not included in the table).}
\vspace{-3mm}
\end{table}

\vspace{-1mm}
\subsection{Simulation}
We trained and tested the policy on workstations with an Intel Xeon 3.6 GHz CPU and an Nvidia Titan GPU. We evaluated the approaches in simulation using the Raisim simulator, and the results are summarized in Table \ref{tab:simulation_results}. The Base Policy in Table \ref{tab:simulation_results} is the policy trained only in one scenario, which means the policy cannot adapt to different terrain and robot configurations. \rev{In Table \ref{tab:simulation_results}, the final fine-tuned AdaptiveON has 15$\%$ improvement in stability and drift reduction than the base policy in uneven terrain and reduces elevation changes in average 20$\%$ at each step in hills scenarios.} We generated three different scenarios: 1. Uneven Terrain with different frictions and high bumpiness; 2. Static Obstacles with some small variations in friction and bumpiness; 3. Hills with high elevation differences from bottom to top. Each of the methods was executed 100 times in each of these scenarios to calculate the following metrics. 

\invis{The scenarios with uneven terrain in the Raisim simulator have different frictions and high bumpiness. }
The \textit{\textbf{vibration}} metric is measured as the average values of vibrations accumulated along the trajectory:

\begin{equation}
\text{vibration}=\frac{1}{N}\sum_{t=0}^N (vb_t),
\label{eqn:stability}
\end{equation} 

Each vibration at time step $t$ is defined in Equation \ref{eqn:single_stability},
 and N is the final time step. Our approach is approximately $5$ times more stable on uneven terrains than the other methods. The Stable Model 
ensures lower vibrations by running the robot slower on the bumpy areas.

From observation, the linear velocity doesn't change much in locomotion on slippery terrain and drifts are mostly caused by turning. Therefore, the \textit{\textbf{drift}} metric is measured by the heading error, which is the difference between the actual direction and the calculated direction, which is calculated by the last direction and the given action.
 

\begin{equation}
    \text{drift} = \frac{1}{N}\sum_{t=0}^N\norm{d_{a, t+1} - d_{c, t+1}(\a_t, d_{a, t})},
    \label{eqn:drift_sim}
\end{equation}

where $p_{a, t+1}$ is the actual position of the next time step $t+1$, and $p_{c, t+1}$ is the calculated next position at $t+1$ based on the velocity and current position. The Adaptive Model can alleviate the drifting issues. From observation, it can drive the robot with lower angular velocity when the robot runs on slippery terrains.

\invis{The scenarios with static obstacles also have some small variations in friction and bumpiness. We can see that our model has a relatively high success rate and is comparable with other methods with effectively high success rates at avoiding static obstacles.}

\invis{For the Hills scenarios, these terrains have a high elevation difference from bottom to top.} 
The \textit{\textbf{elevation changes}} metric is defined as follows:
\begin{equation}
   \text{elevation\_change} = \frac{1}{N} \sum_{t=0}^N \abs{e_{t+1}-e_{t}},
   \label{eqn:elevation_changes}
 \end{equation}
where $e_t$ is the elevation of the robot in the time step $t$.
To highlight the performance for climbing hills, we consider scenarios with significantly low bumpiness and high friction. Then, we compare the algorithms with different start and goal locations. From Table \ref{tab:simulation_results}, the $0.9$ success rate of CrowdSteer is caused by the fact that the algorithm cannot detect elevation changes, and the robot flips over when it runs on higher elevation gradients. Compared to the methods, our algorithm results in relatively lower elevation changes and a higher success rate, although the trajectory is longer than other algorithms.

 As a baseline of navigation, we also tested collision avoidance in dense scenarios with objects 2m apart. AdaptiveOn has a relatively high success rate compared with other approaches (Table~\ref{tab:simulation_results}). Note that we did not include a comparison for Ego-Graph for the Static Obstacles scenario as it does not provide the ability to specifically handle obstacles.

\subsection{Field Experiments}


To demonstrate that our approach can adapt to other differential driving robots, we implement the approach on both Jackal and Husky robots (Figure \ref{fig:frontpage}). Both robots are equipped with Velodyne VLP16 3D Lidar mounted on top of the robot and IMU (we use a Zed camera to get IMU data because it is more accurate and less noisy than the robot’s). The computational platforms are composed of an Intel i9 CPU and an Nvidia RTX 2080 GPU. The running speed of the algorithm is around 5Hz.

\begin{figure*}[htp]
\centering
\includegraphics[width=0.9\linewidth]{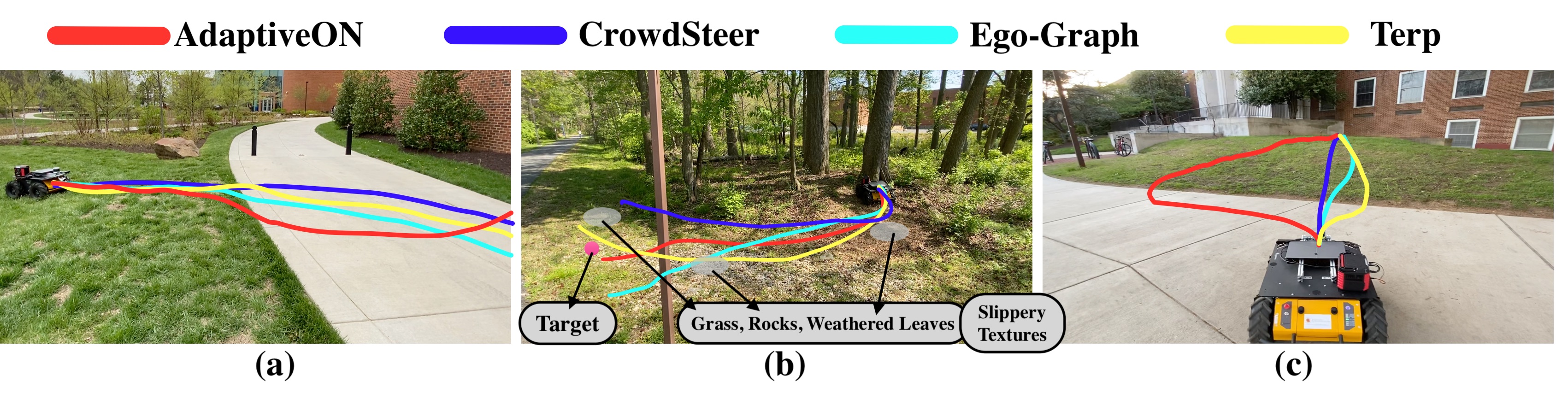}
\caption{Real-world deployment of the Husky robot on different terrains: (a) bumpy terrain with small slope gradient; (b) unstructured slippery terrain (grass, rocks, or weathered leaves), which can cause significant drifts when robots take sharp turns; (c) a hill with a high elevation difference between the bottom and the top. In the last scenario, we can see our method can deviate to the area with a smaller gradient. (Note the trajectories of different methods are marked with different colors.)}
\label{fig:real_scenarios}
\end{figure*}

\begin{figure*}[htp]
\centering
{\includegraphics[width=0.7\linewidth, height=0.2\linewidth]{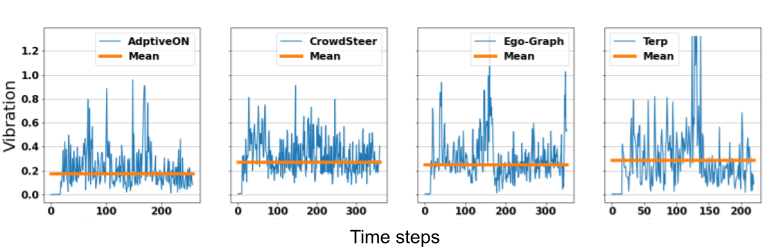}}
\caption{Stability plots of the scenario in Figure \ref{fig:real_scenarios} (a). The lower the value, the less vibration the robot experiences. For all the methods, vibration changes from 0 to 1.4, and our approach has a relatively low vibration without large fluctuations. The peaks occur when the robot runs onto pavement from the grass region.}
\label{fig:stability}
\end{figure*}

The Normal Distributions Transform (NDT)~\cite{ndt} method supports a larger range of initial pose estimates and is faster than the Iterative Closest Points algorithm (ICP)~\cite{compareicpndt}. Thus, for localization and mapping, we use NDT and the Lightweight and Ground-Optimized Lidar Odometry and Mapping (LeGO-LOAM)~\cite{legoloam2018} method, respectively.

\invis{To demonstrate that our approach could be adaptive to other differential driving robots, we implement the approach on both Jackal and Husky robots. As shown in  Figure \ref{fig:frontpage},  the Jackal robot is tested on a slope with an uneven grassy terrain, and the Husky robot is tested on a hill with high elevation changes and on an unstructured slippery area.}

\invis{To evaluate the stability and reliability of the AdaptiveON in real world, we also compared our approach with other approaches in different challenging terrains with some given starts and goal locations.}

\invis{
\subsubsection{Real-world Results}
}

Collision avoidance of static obstacles is the baseline for real-world testing, so we only evaluate the performance of AdaptiveON with other methods (CrowdSteer, Ego-Graph, and Terp) on different challenging terrains with the same starting and goal positions, as shown in Figure \ref{fig:real_scenarios}. \rev{We compared these approaches by taking the average values of 4 runs of each approach.}
The comparison is summarized in Table \ref{tab:realworld_result}, and the results correspond to the trajectory curves in Figure \ref{fig:real_scenarios}. 
All the metrics are measured as they are in simulation. The real-world scenarios mostly merged different properties of terrain, e.g., unevenness, different elevation changes, obstacles, slippery ground, etc. However, each scenario has a different major navigation challenge for the robot.

\begin{table}[htp]
\resizebox{\linewidth}{!}{
\begin{tabular}{|c|c|c|c|c|c|c|c|} 
\hline
\multirow{3}{*}{Scenarios} & \multirow{3}{*}{Methods} & Trajectory & Time  & Vibration & Elevation   & Drift \\ [0.5ex] 
 &  & Length & (s) & (r/step)  & Changes  &  (r/step) \\ [0.5ex] 
 &  &  (m) &  &  $\times$ 100 & (cm/step)  &   $\times$ 100 \\ [0.5ex] 
\hline
\multirow{2}{*}{Uneven} 
& AdaptiveON & 9.56 & 10.8 & 17.39 & 0.05 & 2.16\\
& CrowdSteer & 9.16 & 12.0 & 27.19 & 0.01 & 2.10\\
\multirow{2}{*}{Terrain} 
& Terp       & 9.81 & 11.1 & 29.11 & 0.20 & 6.50\\
& Ego-graph  & 9.48 & 12.0 & 25.10 & 0.12 & 5.33\\
\hline

\multirow{2}{*}{Slippery} 
& AdaptiveON & 5.05 & 10.2 & 24.81 & 0.03 & 3.69 \\
& CrowdSteer & 6.79 & 11.9 & 45.05 & 0.02 & 15.90 \\
\multirow{2}{*}{Terrain} 
& Terp       & 7.69 & 12.1 & 38.70 & 0.03 & 4.01 \\
& Ego-graph  & 8.10 & 8.4 & 32.31 & 0.04 & 12.32 \\
\hline

\multirow{4}{*}{Hills} 
& AdaptiveON & 13.24 & 24.2 & 10.23 & 0.52 & 0.76\\
& CrowdSteer & 11.38 & 14.4 & 22.71 & 0.78 & 1.95\\
& Terp       & 11.13 & 14.5 & 30.34 & 0.61 & 1.84\\
& Ego-graph  & 10.30 & 10.9 & 23.15 & 0.70 & 0.97\\
\hline
\end{tabular}
}
\caption{\label{tab:realworld_result} 
Real-world deployment results: The lower the Vibration, Elevation Changes, and Drift values, the better (more reliable) the performance. Compared to other methods, our approach has at least ${30.7\%}$ improvement in stability on uneven terrain, at least ${8.08\%}$ reduction of drifts on slippery terrain, and at least ${14.75\%}$ improvement on the elevation change in each step on the hills.
}
\end{table}

\textbf{Uneven Terrain:} Scenario (a) tests the stability on uneven terrains, where the grass and the edge of the pavement cause vibrations. The curb of the grassy area is significantly bumpy due to the sudden elevation changes. From the trajectories, our approach and Terp tend to prevent running straight across the edge, while CrowdSteer and Ego-Graph navigate on the concrete area with larger pitch angles. Figure \ref{fig:stability} shows the stability changes on this terrain, where the vibration is measured by Equation \ref{eqn:single_stability}. We observe that our approach results in less vibration than other methods. The last part of the trajectory is on the concrete area, where all the methods have relatively small vibration values. The peaks of the graphs are when the robot runs across the edge of the grass terrain. We can see our method has lower values for and less fluctuation of vibration. In contrast, CrowdSteer operates at high speeds, which makes the average vibration amplitude and variation significantly high. Ego-Graph and Terp tend to choose areas with lower elevation changes, but they depend heavily on waypoints instead of actions, and the vibration values are very high on the grass regions. We observe that the peak values for Terp and Ego-Graph are lower than CrowdSteer; on the edge of the grassy area, CrowdSteer rushes the robot onto the concrete, which causes significantly high pitch angle changes, whereas other methods navigate from a smaller angle with the curb. As shown in Table \ref{tab:realworld_result}, our method has at least $30.7\%$ improvement in stability.

\textbf{Slippery and Unstructured Terrain: } Scenario (b) is a slippery terrain with unstructured features. The rocks are loose and withered leaves are soft. Combined with bumpy grass textures, the scenario can easily lead to drifting when robots move on the terrain.  This scenario also has trees that the robot must avoid. The purple point is the target for all the methods. From the trajectories, we can observe that CrowdSteer deviates most because it moves the robot very fast with sharp turns. AdaptiveON and Terp have similar deviations.
As shown in Table \ref{tab:realworld_result}, our approach reduces the drifting at least by $8.08\%$.

\textbf{Hills: } Scenario (c) contains a hilly region with high elevation changes and inconsistent elevation gradients. Further, the grass surface is bumpy and slightly slippery for the robot. We observe that AdaptiveON can navigate the robot along trajectories with lower elevation changes and lower vibrations. Although Terp can also choose lower elevations, its trajectory optimization could lead to relatively higher elevations when the waypoints are significantly far away from the robot.  Hence, as presented in Table \ref{tab:realworld_result}, we can see that our method provides at least a $14.75\%$ decrease in the elevation changes when navigating in hilly regions.

\section{CONCLUSION}
\label{section:conclusions}
We presented AdaptiveON - a reinforcement learning-based method with a multi-stage training pipeline for stable and reliable outdoor navigation. We showed the performance on complex outdoor terrains through both simulation and real-world deployments. Our method is modular and can be used as a baseline for adding other functionalities. It is possible to add new models using similar architecture and training strategies to achieve different functions.

Nevertheless, there are still some limitations. 
The stable model depends on the vibration of the last three steps, which is latent and causes pick values of vibrations when the robot runs on a bumpy area in the current step. In addition, there are still many other complex scenarios our approach could not solve. e.g., crossing the area with tall grass, staying away from cliffs, or detecting fast-moving obstacles. 

In the future, this work can be improved in several respects: To prevent the interference between preventing drifting and making robots run stably, those two functions can be decoupled, and the robot should prioritize using functions in different situations. More perceptive features can be used to predict the bumpiness of the terrain to reduce the extreme values. \rev{It can be integrated with a high level path planning algorithms that generates waypoints, to ensure more stable and reliable navigation between waypoints.}


%
\IEEEpeerreviewmaketitle

\appendices

\begin{figure*}[!ht]
    \centering
    \includegraphics[width=0.9\linewidth]{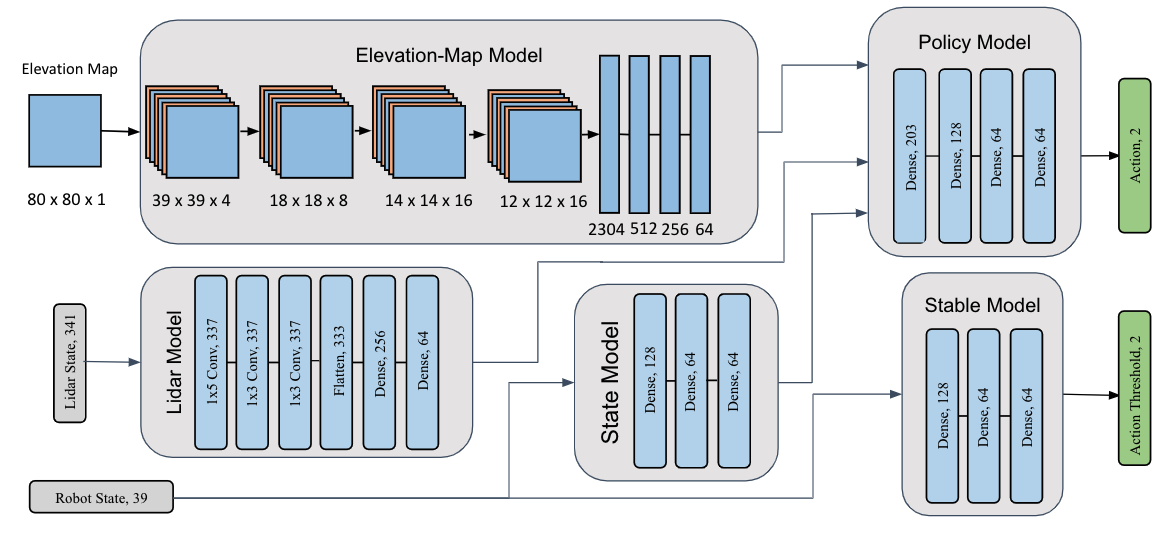}
    \caption{Details of the different models}
    \label{fig:allmodels}
\end{figure*}

\begin{figure*}[!htp]
\minipage{0.2\textwidth}
  \includegraphics[width=\linewidth, height=0.8\linewidth]{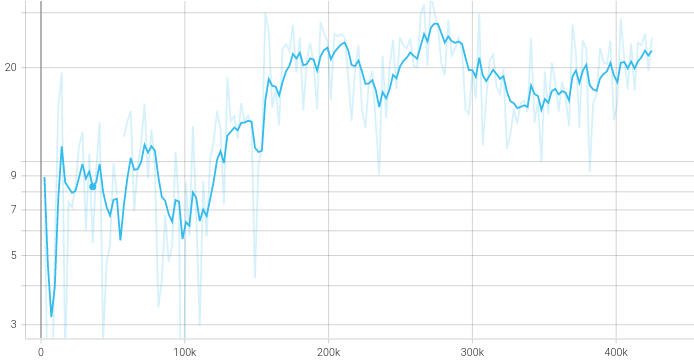} 
  \centering{Stage 1}
\endminipage\hfill
\minipage{0.2\textwidth}
  \includegraphics[width=\linewidth, height=0.8\linewidth]{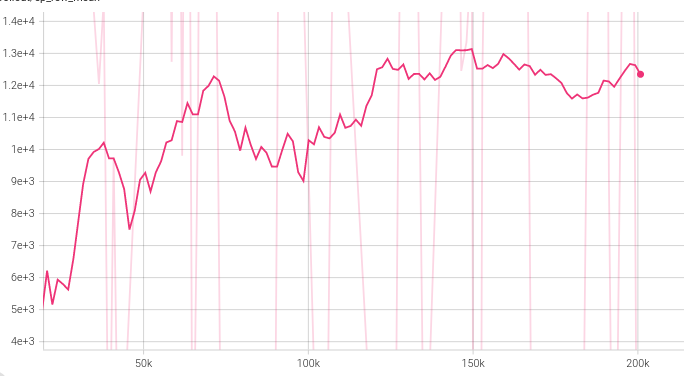} 
  \centering{Stage 2}
\endminipage\hfill
\minipage{0.2\textwidth}%
  \includegraphics[width=\linewidth, height=0.8\linewidth]{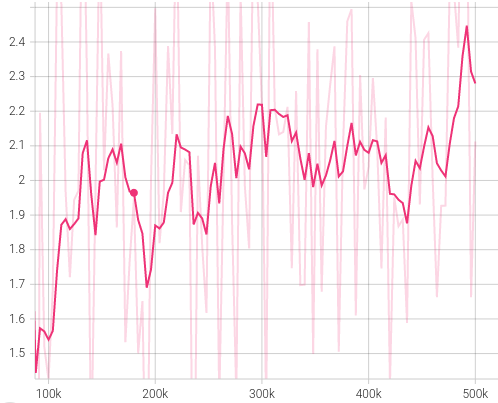}
  \centering{Stage 3}
\endminipage\hfill
\minipage{0.2\textwidth}%
  \includegraphics[width=\linewidth, height=0.8\linewidth]{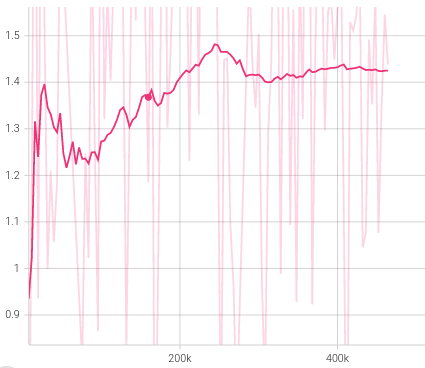}
  \centering{Stage 4}
\endminipage\hfill
\minipage{0.2\textwidth}%
  \includegraphics[width=\linewidth, height=0.8\linewidth]{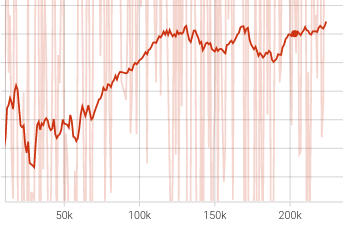}
  \centering{Stage 5}
\endminipage
\caption{The figures are from the training of different functional stages. The trainings all converged well.}
\label{fig:training_curve}
\end{figure*}

\section{Networks}
The input of the Lidar Model is a single-channel Lidar data, which is also normalized to $[0,1]$ by a 10m Lidar range. The input of the Elevation-Map Model is an elevation map generated from the model~\cite{elemap} with the raw Lidar data as its input. The elevation map is also normalized by first being clipped by the height range $[-2m, 2m]$ and then being divided by 2m. The Lidar Model extracts
the nearby obstacles, and the Elevation-Map model 
extracts the nearby terrain. The elevation map is an ego-centric grid map with the shape of $(80 \times 80)$. To improve the training convergence, we normalized the elevation map by the maximum and minimum elevations $(-2\text{m}, 2\text{m})$. This feature is encoded by the Elevation-map Model, as in Figure \ref{fig:allmodels}.


The Lidar data is processed by the Lidar Model, which is the other perceptive model. The input Lidar data is normalized by a range of 10m. We use 1-D convolutional layers and fully-connected layers to process the Lidar data. The Lidar Model is as in Figure \ref{fig:allmodels}.

The State Model processes state data by several fully-connected layers, as in Figure \ref{fig:allmodels}.

The Policy Model and the Stable Model are composed of fully-connected layers. They outputs are the action and the threshold of the action, as in Figure \ref{fig:allmodels}.

For different training stages, we use different total time steps, and the training convergences are as in Figure \ref{fig:training_curve}. The training stages run sequentially, and we load the models from the last stage in each stage. The rewards also converged to different values according to different reward functions.

\section{Experiments}
\begin{figure*}
    \centering
    \includegraphics[width=0.8\linewidth]{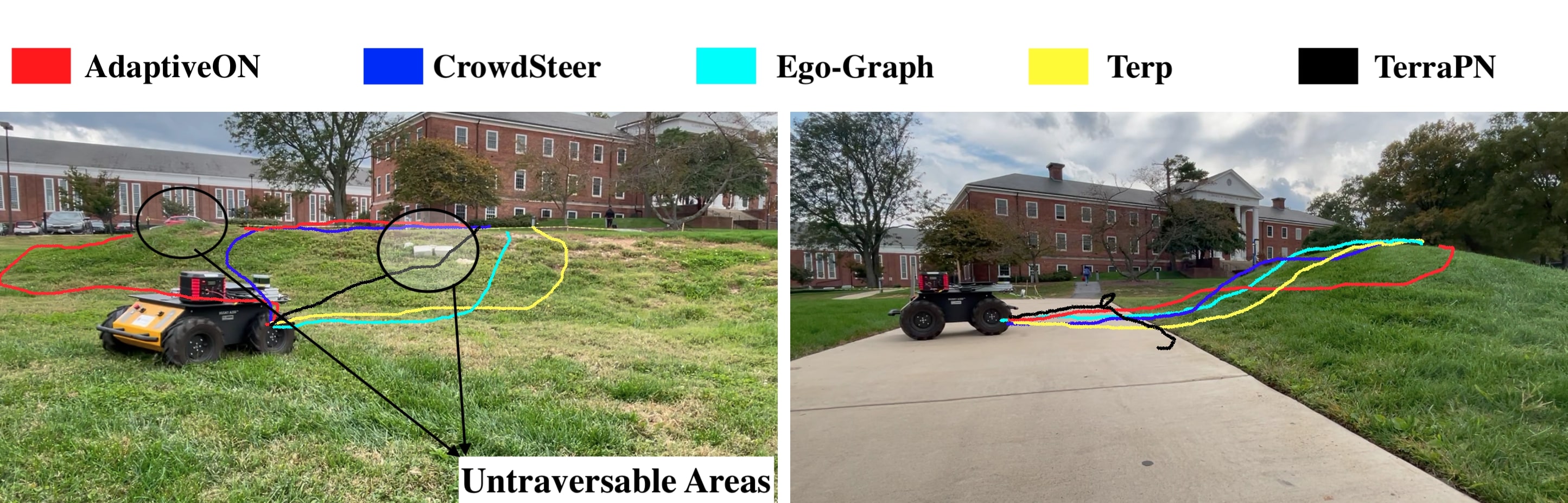}
    \caption{These two scenarios include an untraversable area (left), where the robot is easily flipped over, and a dangerous area (right). In these two scenarios, the robot needs to go up a high hill. We can see our approach (red) takes the robot to the top of the hill with lower elevation changes.}
    \label{fig:sup_scenarios}
\end{figure*}

\rev{To further demonstrate the capabilities of our model, we conduct more experiments in challenging areas and compare the results with another approach, TerraPN~\cite{terrapn}, in these two scenarios. In Figure \ref{fig:sup_scenarios}, all the hills have heights of more than 1.5m. Scenario (a) has some dangerous areas with sharp gradients (shown in black circles in Figure \ref{fig:sup_scenarios}). The white concrete area is particularly dangerous because the robot could be damaged running on the area. Scenario (b) is a hill with an average high gradient in all the directions. In this scenario, Ego-graph takes lots of time to recalculate the best trajectory, meaning it has the slowest speed. From the Table \ref{tab:sup_scenarios}, we can observe that our method has less vibration and lower elevation changes while moving on different challenging terrains. In Scenario (a) we also have less drift than other approaches. In Scenario (b) CrowdSteer has less drift because it doesn’t move around the hill. Other methods move around the hill to reduce the unstable sharp elevation changes.
} 

\begin{table}[htp]
\resizebox{\linewidth}{!}{
\begin{tabular}{|c|c|c|c|c|c|c|c|} 
\hline
\multirow{3}{*}{Scenarios} & \multirow{3}{*}{Methods} & Trajectory & Time  & Vibration & Elevation   & Drift \\ [0.5ex] 
 &  & Length & (s) & (r/step)  & Changes  &  (r/step) \\ [0.5ex] 
 &  &  (m) &  &  $\times$ 100 & (cm/step)  &   $\times$ 100 \\ [0.5ex] 
\hline
\multirow{5}{*}{Scenario (a)} 
& AdaptiveON & 58.27 & 49 & 30.44 & 1.09 & 4.03\\
& CrowdSteer & 26.77 & 32 & 40.93 & 6.45 & 5.03\\
& Ego-graph  & 17.18 & 19 & 66.30 & 5.95 & 7.07\\
& Terp       & 21.62 & 22 & 58.06 & 6.39 & 9.01\\
& TerraPN    & 9.43  & -  & 51.08 & 10.02 & -\\
\hline

\multirow{5}{*}{Scenario (b)} 
& AdaptiveON & 18.15 & 22 & 45.23 & 1.18 & 10.43\\
& CrowdSteer & 12.90 & 8 & 33.13 & 5.20 & 7.81\\
& Ego-graph  & 11.44 & 39 & 47.81 & 2.57 & 18.27\\
& Terp       & 10.70 & 8 & 44.84 & 1.90 & 9.97\\
& TerraPN    & - & - & - & - & -\\
\hline
\end{tabular}
}
\caption{\label{tab:sup_scenarios} The scenarios (a) and (b) are the scenarios from Figure \ref{fig:sup_scenarios}. Our method has better stability and fewer elevation changes in motion. We also observe that our approach has less drift in Scenario (a). TerraPN cannot run on Scenario (b) because it treats the grass hill as a forbidden area. It cannot avoid the dangerous area of the Scenario (a) either, so the odometry-calculated trajectory goes wrong when the robot runs into the dangerous area. Therefore, we cannot calculate the drift value. 
}
\vspace*{-5mm}
\end{table}

\section*{Acknowledgment}
This work was supported in part by ARO Grant, W911NF2110026  and U.S. Army Cooperative Agreement W911NF2120076

\ifCLASSOPTIONcaptionsoff
  \newpage
\fi

\bibliographystyle{IEEEtran}
\bibliography{references}

\end{document}